\documentclass[10pt,twocolumn,letterpaper]{article}

\usepackage{color}
\usepackage{caption}
\usepackage[pagenumbers]{iccv} %

\definecolor{iccvblue}{rgb}{0.21,0.49,0.74}
\usepackage[pagebackref,breaklinks,colorlinks,allcolors=iccvblue]{hyperref}

\title{Controllable and Expressive One-Shot Video Head Swapping}

\author{
    Chaonan Ji \quad Jinwei Qi \quad Peng Zhang \quad Bang Zhang \quad Liefeng Bo \\
    Tongyi Lab, Alibaba Group  \\
    \tt\small \url{https://humanaigc.github.io/SwapAnyHead/}
}

\begin{document}
\twocolumn[{
    \renewcommand\twocolumn[1][]{#1}
    \maketitle
    \vspace*{-2.9em}
    \begin{center}
        \captionsetup{type=figure}
        \includegraphics[width=1.0\linewidth]{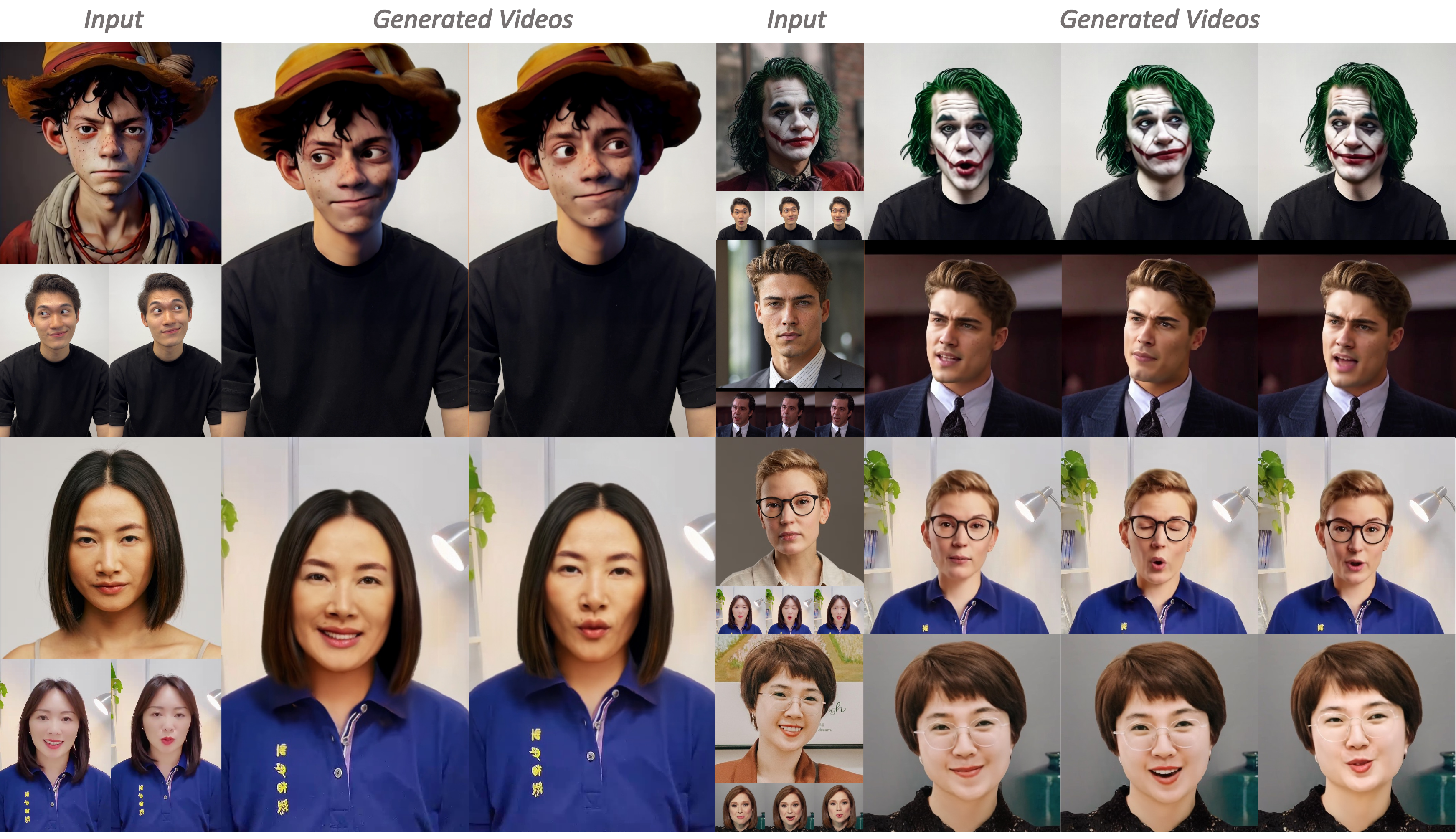} \vspace{-0.8em}
        \captionof{figure}{
            \textbf{Qualitative results of our method.} Given a reference image and video sequence as input, our model can generate high-fidelity head swapping results that accommodate diverse hairstyles, expressions, and identities.
        }
        \label{fig:overview}
    \end{center}
}]
\begin{abstract}

In this paper, we propose a novel diffusion-based multi-condition controllable framework for video head swapping, which seamlessly transplants a human head from a static image into a dynamic video, while preserving the original body and background of target video, and further allowing to tweak head expressions and movements during swapping as needed. Existing face-swapping methods mainly focus on localized facial replacement neglecting holistic head morphology, while head-swapping approaches struggling with hairstyle diversity and complex backgrounds, and none of these methods allow users to modify the transplanted head’s expressions after swapping. To tackle these challenges, our method incorporates several innovative strategies through a unified latent diffusion paradigm. 1) Identity-preserving context fusion: We propose a shape-agnostic mask strategy to explicitly disentangle foreground head identity features from background/body contexts, combining hair enhancement strategy to achieve robust holistic head identity preservation across diverse hair types and complex backgrounds. 2) Expression-aware landmark retargeting and editing: We propose a disentangled 3DMM-driven retargeting module that decouples identity, expression, and head poses, minimizing the impact of original expressions in input images and supporting expression editing. While a scale-aware retargeting strategy is further employed to minimize cross-identity expression distortion for higher transfer precision. Experimental results demonstrate that our method excels in seamless background integration while preserving the identity of the source portrait, as well as showcasing superior expression transfer capabilities applicable to both real and virtual characters.

\end{abstract}    
\section{Introduction}
\label{sec:intro}

Video head swapping, in particular, seeks to seamlessly replace a source head in a target video while preserving the source individual's identity and target video's background and expression, offering immense application potential in film production, virtual reality and advertisement composition.

Face swapping has long been a topic of interest. Most face-swapping methods \cite{simswap,DBLP:conf/cvpr/KimLZ22,DBLP:journals/corr/abs-1912-13457,DBLP:journals/pami/NirkinKH23,DBLP:conf/ijcai/0002CZCTWLWHJ21,DBLP:conf/cvpr/XuDWJPH22,DBLP:conf/cvpr/LiL0SH21}  aim at replacing localized facial regions by injecting identity features, yet they struggle to effectively transfer hairstyles, head shapes. In contrast, Video head swapping that focus on replacing the entire head in the target image, has been relatively less explored. StylePoseGAN \cite{DBLP:journals/corr/abs-2102-11263} employs GAN inversion for head swapping but faces challenges in maintaining consistency of body and background; HeSer \cite{heser} utilizes two-stage training to decouple facial expressions and merge backgrounds, but it struggles with transferring hairstyles and inpainting complex backgrounds, often resulting in noticeable artifacts. Certain Diffusion-Based methodologies \cite{DBLP:journals/corr/abs-2308-06721,DBLP:conf/cvpr/RuizLJPRA23} integrate identity features via self-attention mechanisms, which effectively facilitate the retention of identity characteristics. However, such methods are unable to generate images under specified background and body conditions.

Moreover, to ensure seamless head-swapping in video sequences, accurate and expressive facial expression transfer is indispensable.
IPAdapter \cite{DBLP:journals/corr/abs-2308-06721}, DreamBooth \cite{DBLP:conf/cvpr/RuizLJPRA23} and HS-Diffusion \cite{wang2023hsdiffusionsemanticmixingdiffusionhead} lacks the capability to control the facial expressions in the generated images. In contrast, 
AniPortrait \cite{aniportrait} can transfer target expressions but is limited by the lack of expressiveness in facial templates. Follow-your-emoji \cite{emoji} employs 3D landmarks for more expressive transfer but the results may be influenced by the reference image's original expressions. Expression transfer accuracy in both methods declines when there are significant differences in the proportional sizes of facial features between the source and target images. Neither method is adequate for head swapping.

In summary, current head swapping methods encounter two principal challenges: 1) preserving the identity of the source image (notably head shape and hairstyles) while achieving seamless integration with the target image's background; 2) accurately transferring expressions from the target video without being influenced by the facial proportions and expressions of the source image.

In this paper, we propose a controllable one-shot diffusion-based framework for video head swapping, that ensures high-quality background generation and identity consistency while allowing for expressive and controllable transferring of facial expressions and facilitating natural head movements. 
We reformulate head swapping as a conditional inpainting task for the head region, guided by identity features, background cues, and 3D facial landmarks. After segmenting and removing the head region of the target image, our model generates a new head to seamlessly integrate into the designated region while maintaining consistency with the surrounding background and body.

However, head segmentation may lead to shape leakage issues. To counter this, we introduce a shape-agnostic mask strategy designed to minimize head shape leakage issues by disrupting the segmented boundaries of the head. In addition, segmentation naturally creates a link between the hair and the body, particularly with long hair resting on the shoulders. We propose a hair enhancement strategy that disrupts the relationship between the hair and body, enabling the generated results to retain the hairstyle from the source image, supporting head swapping across diverse hairstyles. 

Additionally, 
we introduce an expression-aware landmark retargeting module that decouples identity, expression, and pose, allowing for expression transfer and editing. This module eliminates expressions from the source image and adjusts driving landmarks according to the proportions of facial features. Specifically, we first employ a 3DMM \cite{faceverse} to fit the source image, obtaining a neutral expression template through expression coefficients adjustments. Subsequently, we determine the source landamrks by aligning the adjusted facial template's projected landmarks with 3D landmarks directly detected by the Mediapipe \cite{mediapipe}. This process yields the source 3D landmarks in neutral expression, thus maintaining the expressive capacity of the Mediapipe landmarks while effectively removing the original expressions present in the source image. Building on this, we further adjust the expression intensity according to the size ratios of the facial features between the source and target individuals, which enhances the accuracy of the expression transfer and facilitating expression editing. 

To conclude, our main contributions are the following:
\begin{itemize}
\item A novel controllable one-shot diffusion-based framework for video head swapping that ensures high-quality background generation and preserves identity while allowing for expressive facial expression transfer and editing.
\item We propose a shape-agnostic mask strategy alongside a hair enhancement strategy that better preserves identity and supports diverse hairstyles in head swapping.
\item We present an expression-aware landmark retargeting module that preserves expressive facial driving, removes the source image's expression influence, and accommodates expression transfer across portraits with different facial feature proportions.
\end{itemize}

\section{Related Work}
\label{sec:relatedwork}

\subsection{Face and Head Swapping}
Face swapping \cite{simswap,DBLP:conf/cvpr/KimLZ22,DBLP:journals/corr/abs-1912-13457,DBLP:journals/pami/NirkinKH23,DBLP:conf/ijcai/0002CZCTWLWHJ21,DBLP:conf/cvpr/XuDWJPH22,DBLP:conf/cvpr/LiL0SH21,diffswap} has become a prominent research topic, aimed at fusing the identity information extracted from the source face to the target face. However, such methods typically do not adequately consider the head shape and hairstyle, which constrains the similarity of the swapped face to the source image. In contrast, head swapping \cite{heser,deepface,DBLP:journals/corr/abs-2102-11263,insetgan,wang2023hsdiffusionsemanticmixingdiffusionhead,mat} that focuses on exchanging the entire head region, remains comparatively underexplored.

Deepfacelab \cite{deepface} attempts to address the head swapping challenge by utilizing multiple reference images; however, the swapped results encounter artifacts in the fusion regions. GAN-based approaches \cite{insetgan,DBLP:journals/corr/abs-2102-11263,mat} enable head swapping by editing localized face areas, yet often struggle with maintaining background consistency and exhibit difficulties in generalization. HeSer \cite{heser} integrates head driving with background fusion algorithms, achieving high-quality head swapping. However, it faces limitations in generating diverse hairstyles, transferring complex expressions and maintaining background. IPAdapter and related approaches \cite{DBLP:journals/corr/abs-2308-06721,DBLP:conf/mm/FanYLZZ24,DBLP:conf/cvpr/RuizLJPRA23} seek to incorporate head identity information and employ text prompts to generate diverse, identity-consistent images. However, these methods face challenges in precisely controlling backgrounds and expressions. HS-Diffusion \cite{wang2023hsdiffusionsemanticmixingdiffusionhead} applies a semantic-mixing diffusion model guided by human parsing for head swapping, though it is unable to effectively transfer expressions. In response to these challenges, we propose a head swapping framework based on Latent Diffusion Models (LDMs) \cite{RombachBLEO22} that successfully integrates target background, diverse hairstyle, and expression transfer, facilitating high-quality video head swapping.

\subsection{Expression transfer}
GAN-based expression transfer has attracted substantial research interest over time. Previous methods \cite{wav2lip,obama,styleSync,SyncTalkFace,SongWQHL22,SongZLWQ19,ThiesETTN20,ChenLMDX18,ZhangLGH24} have primarily concentrated on lip-sync synthesis within the mouth region. To enhance the overall expressiveness of talking face, many methods \cite{GuoCLLBZ21,JiZWWWXC22,LiangPGZHHHLD022,SunZ0K21,Wu0WDDD21,WangM021,PangZQFCS023,WangZSZ0YL23,TanJP24,LiuC0DC0024,ZhangQZZW0CW023,EMOPortraits,VASA,float,liveportrait,lia,megaportrait,sadtalker} attempt to encompass the entire head region.

Significant efforts have been dedicated to refining these control signals to enhance the precision of expression control. StyleAvatar \cite{WangZSZ0YL23} employs 3D Morphable Model (3DMM) \cite{faceverse} for facial expression transfer. Face vid2vid \cite{WangM021}, SadTalker \cite{sadtalker} along with LivePortrait \cite{liveportrait} utilize implicit keypoints to achieve high-quality expression transfer. Furthermore, MegaPortrait \cite{megaportrait}, EMOPortrait \cite{EMOPortraits}, VISA \cite{VASA}, and LIA \cite{lia} adopts motion latent derived from images as control signals, which effectively bypass the constraints of 3DMM but introduce the risk of identity leakage. While these techniques display commendable expressiveness, they are inherently limited by the generative capacities of GANs, which often hinders their ability to faithfully transfer expressions.

Recently, Latent Diffusion Models (LDM) \cite{RombachBLEO22} have demonstrated superior generative capabilities, achieving high-quality video generation via text \cite{he2023latentvideodiffusionmodels,SingerPH00ZHYAG23,DBLP:conf/iclr/0002YRL00AL024,DBLP:journals/corr/abs-2310-19512,DBLP:journals/corr/abs-2211-11018} or speech inputs \cite{TianWZB24}. However, the weak control signals pose challenges for precise expression control, thus many methods explore spatial-aligned control signals \cite{MaHCWC0C24,DBLP:journals/corr/abs-2403-08268,DBLP:journals/tvcg/XingXLZZHLCCWSW25, magicpose, magicanimate, DBLP:conf/iclr/Zhang0J0Z024, aniportrait, emoji, xportrait} by ControlNet \cite{ZhangRA23} or PoseGuider \cite{aa}. MagicAnimate \cite{magicanimate} employs densepose map as control signal, successfully transferring poses but struggling to accurately replicate expressions. AniPortrait \cite{aniportrait} and MagicDance \cite{magicpose} utilizing facial landmarks derived from 3DMM, facilitating high-quality cross-identity image generation, and Follow-you-emoji \cite{emoji} advances this methodology by incorporating 3D landmarks detected by Mediapipe \cite{mediapipe}, enabling more expressive expression transfers.
Our approach builds upon these advancements by integrating 3DMM template into the 3D landmarks derived from Mediapipe \cite{mediapipe}, thereby achieving more precise expression transfer while maintaining identity consistency.

\section{Method}
\label{sec:method}

\begin{figure*}[htb]
    \centering
    \includegraphics[width=\textwidth]{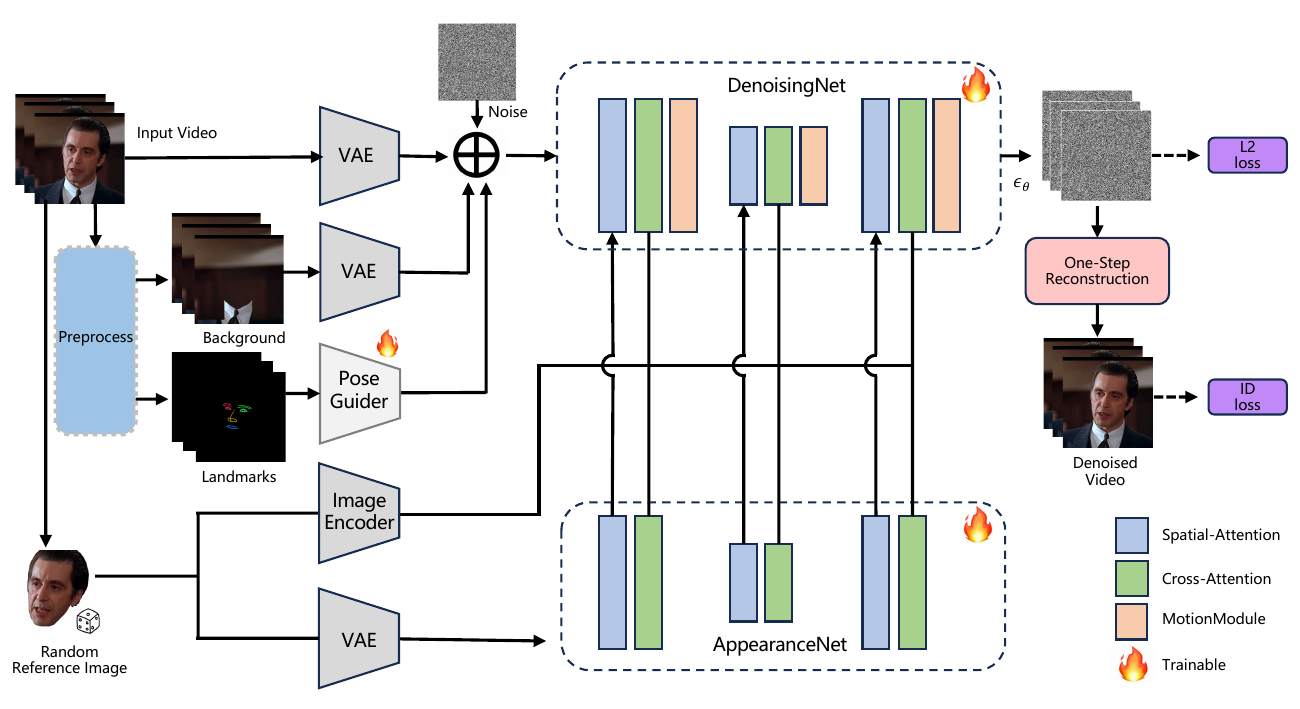}
    \caption{The framwork of our method. During the training stage, we begin by preprocessing the input video using the method outlined in Sec.\ref{medthod:shape} to obtain the inpainted background and detect 3D landmarks uiltizing Mediapipe \cite{mediapipe}. Both the inpainted background and the 3D landmarks serve as conditional inputs, and a frame is randomly selected as the reference image. We reconstruct the image from output noise and introducing additional ID losses in pixel level to enhance identity consistency.}
    \label{fig:pipeline}
\end{figure*}

\subsection{Overview}
In this section, we first present our framework in Sec.\ref{method:framework}. We then detail the Shape-Agnostic Mask and Hair Enhancement strategy in Sec.\ref{medthod:shape}. Lastly, we introduce the Expression-Aware Landmark Retargeting in Sec.\ref{method:expression}. The pipeline is illustrated in Fig.\ref{fig:pipeline}.

\begin{figure}[t]
    \centering
    \includegraphics[width=\columnwidth]{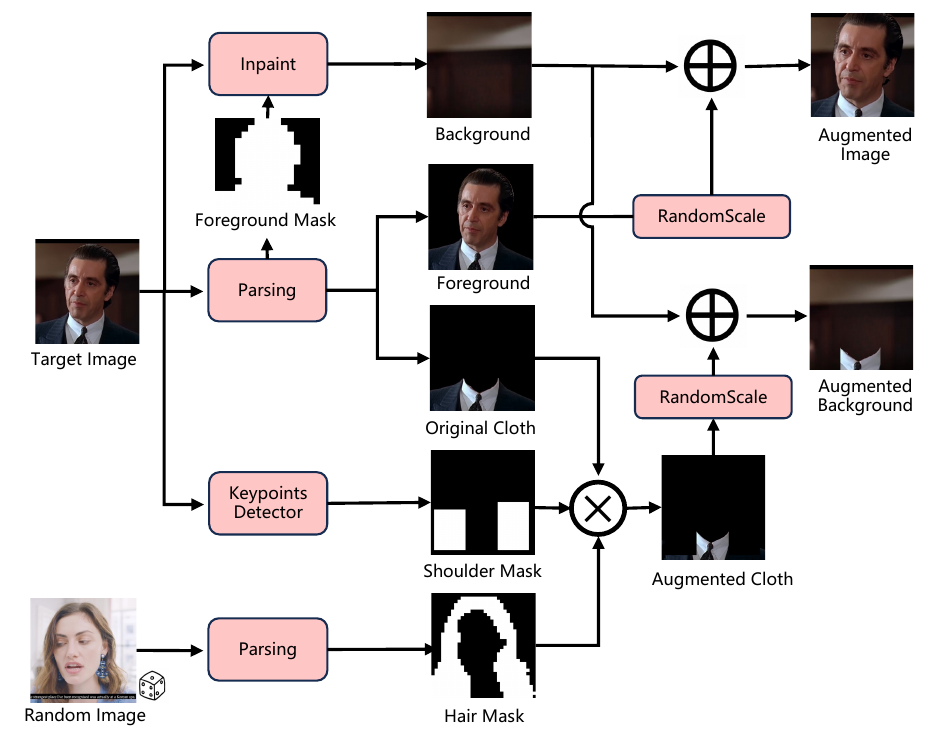}
    \caption{The detail of Shape-Agnostic Mask and Hair Enhancement Strategy. Given a target image, we first obtain the foreground and cloth masks. Then we apply a Shape-Agnostic Mask Strategy to disrupt the segmentation boundaries and obtain an inpainted background. The foreground is randomly scaled and combined with this background to create the augmented image used as training data. Additionally, a random long-haired image is selected to extract a hair mask, and a shoulder mask is created using shoulder keypoints detected by Mediapipe. The original cloth, shoulder mask, and hair mask are combined as the SD input.
    }
    \label{fig:method_shape}
\end{figure}

\subsection{Framework}
\label{method:framework}
Our framework is developed based on LDM \cite{RombachBLEO22} and is extended with background and expression conditions. The framework consists of the following modules: 1) \textbf{AppearanceNet:} It extracts identity information from the source image and injects it layer by layer into the DenoisingNet of the Stable Diffusion (SD) model through self-attention blocks, sharing the same architecture as DenoisingNet. 2) \textbf{PoseGuider:} It extracts multi-scale motion features and fuses them with the SD Unet features, establishing a spatial connection between the control signals and the generated image. 3) \textbf{MotionModule:} maintains temporal coherence and consistency across different frames. 4) \textbf{Image Encoder:} replaces the text encoder by encoding the reference image into a one-dimensional feature and injects it into the network through cross-attention blocks, providing additional global information from the reference image. The detail is shown in Fig.\ref{fig:pipeline}

During training, given a reference image $I_{ref}$ and a input video $I_{d}^{1:N}$ where N denotes the number of frames, we preprocess the input video to obtain the background sequence ${I_{b}^{1:N}}$ and extract driving landmarks ${I_{m}^{1:N}}$ to serve as the driving signal. For simplicity , we omit the superscript $N$. The training objective of SD can be written as:
\begin{equation}
    \begin{aligned}
        \mathcal{L}_{LDM} = \underset{t,z_{0},\epsilon }{\mathbb{E}}  \left [ \left \| \epsilon -\epsilon_{\theta }\left ( \sqrt{\bar{\alpha _{t} } }z_{0}+ \sqrt{1 -\bar{\alpha _{t} }}\epsilon ,c,t \right )   \right \|^{2}  \right ] 
    \end{aligned}
\end{equation}
where $z_{0}$ denotes the latent embedding of $I_{d}$, $\epsilon_{\theta }$ is the predicted noise by SD and $\epsilon$ is the ground truth noise at timestep $t$. $c$ is the condition embedding, which represents $c=(I_{b},I_{m},I_{ref})$ for our method. 

To better preserve the identity, we recover the latent embeddings from the predicted noise $\epsilon_{\theta }$ and decode it into an denoised video sequence $\hat{I_{d}}$ using VAE decoder. The detail of the ID loss will be explained in the supplementary materials. Subsequently, we compute the ID loss between the denoised video and the input video:
\begin{equation}
    \begin{aligned}
        \mathcal{L}_{id} &= \mathbb{E}\left [ \left \| \mathcal{M}_{head} \odot (I_{d}-\hat{I_{d}} )  \right \|^{2} \right ] \\&\quad + \mathbb{E}\left [1-CosSim(\mathcal{\varepsilon } _{id}(I_{d}), \mathcal{\varepsilon } _{id}(\hat{I_{d}}) )  \right ]  
    \end{aligned}
\end{equation}
where $\mathcal{M}_{head}$ is the head region mask and $\mathcal{\varepsilon } _{id}$ is a face recognition model \cite{arcface} that encodes a face image into an identity embedding. Our total loss can be formulated as:
\begin{equation}
    \begin{aligned}
        \mathcal{L} = \mathcal{L}_{LDM} + \mathcal{L}_{id}
    \end{aligned}
\end{equation}

During inference, given a reference image and a driving video, our method can generate high-quality head wrapping video with consistent background and expressive expression. 

\noindent\textbf{Condition Images Preprocessing} To eliminate the influence of the background and body from the reference image, we use a human parsing algorithm \cite{khirodkar2024sapiens} to retain only the reference image's head region as the input $I_{ref}$ for the AppearanceNet. Additionally, we randomly scale and rotate the reference image to disrupt the positional relationship between the head and body, preventing identity leakage. In addition, we segment the foreground and clothing portions of the reference image, then employ an inpainting model \cite{DBLP:conf/wacv/SuvorovLMRASKGP22} to inpaint the complete background. The inpainted background is recombined with the clothing portion to form a new background image $I_{b}$, which serves as the input to the network. The landmarks are detected by Mediapipe \cite{mediapipe}.

\begin{figure}[t]
    \centering
    \includegraphics[width=\columnwidth]{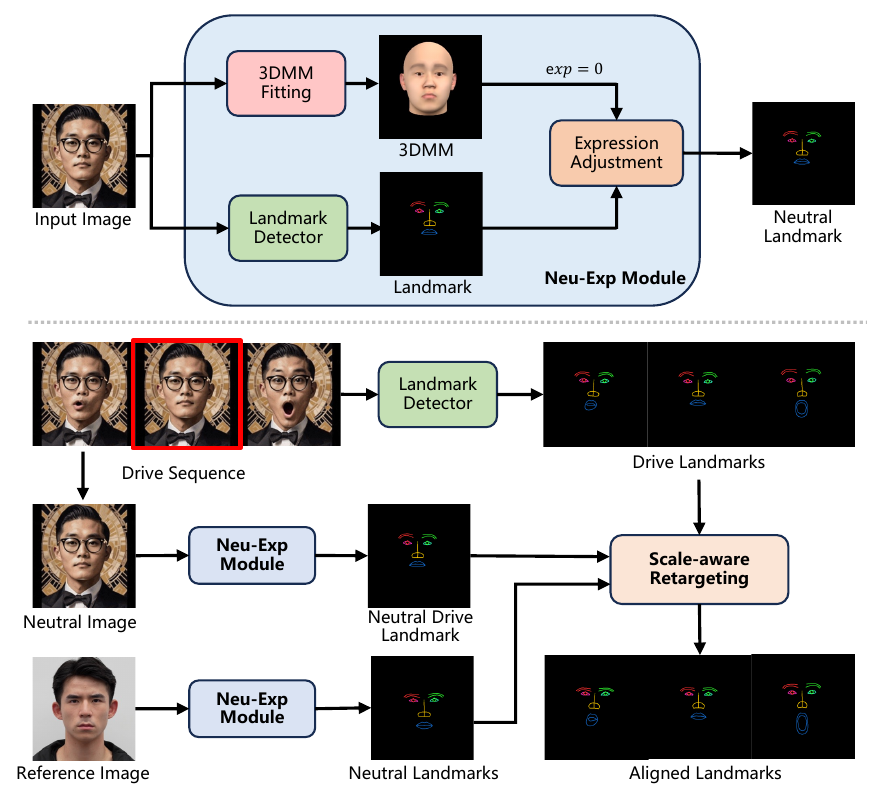}
    \caption{The detail of Expression-Aware Landmark Retargeting. The Neu-Exp module uses a 3DMM template to neutralize the expression in the input image, outputting neutral landmarks. For expression transfer, we select a nearly neutral frame from the drive sequence and process it with the Neu-Exp module to obtain Neutral drive landmarks. Then we use scale-aware retargeting to generate aligned landmarks.}
    \label{fig:method_expression}
\end{figure}

\subsection{Shape-Agnostic Mask and Hair Enhancement}
\label{medthod:shape}

While the the aforementioned framework effectively produces head-swapping results, challenges remain in preserving identity of the reference image and generating diverse hairstyles. These issues are attributed to identity leakage in the training data. Specially, due to the absence of authentic head-swapping videos, we utilize different frames from the same video as the reference image and the target image for training. Both the inputs $I_{ref},I_{b}$ of SD and target video $I_{d}$ share the same identity. The inpainted backgrounds inadvertently encode head shape information through invisible segmentation boundaries (indistinguishable to the human eye but discernible by SD). In addition, this inherently introduce a bias that the generated head is constrained to remain within invisible boundaries, restricting hairstyle diversity.

To address these challenges, we propose a novel Shape-Agnostic Mask strategy. The details are shown in Fig.\ref{fig:method_shape}. Given an input image $I_{d}$, we first obtain the corresponding foreground mask and then dilate it to obtain $M_{f}$. Subsequently, we define a new mask $M_{f}^{new}$, which shares the same dimensions as $M_{f}$ and is composed of non-overlapping blocks of size $(k_{h} \times k_{w})$. Each block in $M_{f}^{new}$ is assigned a uniform value of 0 or 1, determined by the corresponding pixel values in $M_{f}$:
\begin{equation}
    \begin{aligned}
        M_{f}^{new}(i,j) = \begin{Bmatrix}
        1, if \:\:\:any(M_{f}(i,j))>0
         \\
        \:\:\:0, if \:\:\:every(M_{f}(i,j))=0
        \end{Bmatrix} 
    \end{aligned}
\end{equation}
where $M_{f}^{new}(i,j)$ represents value of the block at position $(i,j)$. The $M_{f}^{new}(i,j)$ is used to generate the inpainted background. Since the edges are disrupted, the shape information of the reference image will not be leaked. In addition, during the training phase, we randomly scale the foreground to overcome the bias imposed by the inpainting model, allowing the generated head to extend beyond the limits of $M_{f}^{new}$. Specifically, we first preprocess the input video to obtain the foreground and inpainted background. The foreground is then randomly scaled and recomposed with the inpainted background to produce new training data. This strategy ensures that the foreground has a certain probability of covering areas beyond the original region, thus breaking through the invisible boundaries of the inpainted background. Consequently, during the inference phase, the SD can disregard the influence of invisible boundaries on hair shape and hairstyle.

Furthermore, the shoulder region often provides strong prior information about hairstyles. For example, in cases of long hair draping over shoulders, the cloth segmentation may create holes in the shoulder area reflecting the hair shape from the target image, introducing hairstyle priors during training. During testing, this leads to the generation of long hair in these gaps, even if the reference image depicts a short-haired individual. To address this issue, we propose a Hairstyle Enhancement Strategy, which involves 1) Collecting a set of long-hair videos, and randomly choosing a portrait image $I_{hair}$ from this set during training. Then we align its facial features with the target image, and obtain a hair region mask $M_{hair}$. 2) Using shoulder keypoints detected by Mediapipe, and generating random rectangular masks $M_{rect}$ around these keypoints. 3) Eroding the original cloth region  with $M_{hair}$ and $M_{rect}$ to disrupt hair prior. The final cloth mask can be represented as:

\begin{equation}
    \begin{aligned}
        M_{cloth}^{new} = M_{cloth}\odot (1-M_{hair})\odot (1-M_{rect})
    \end{aligned}
\end{equation}
where $M_{cloth}$ is original body mask and $M_{cloth}^{new}$ is the augmented body mask. This approach effectively mitigates undesired hairstyle priors in the shoulder area.

\subsection{Expression-Aware Landmark Retargeting}
\label{method:expression}

\begin{figure*}[htb]
    \centering
    \includegraphics[width=\textwidth]{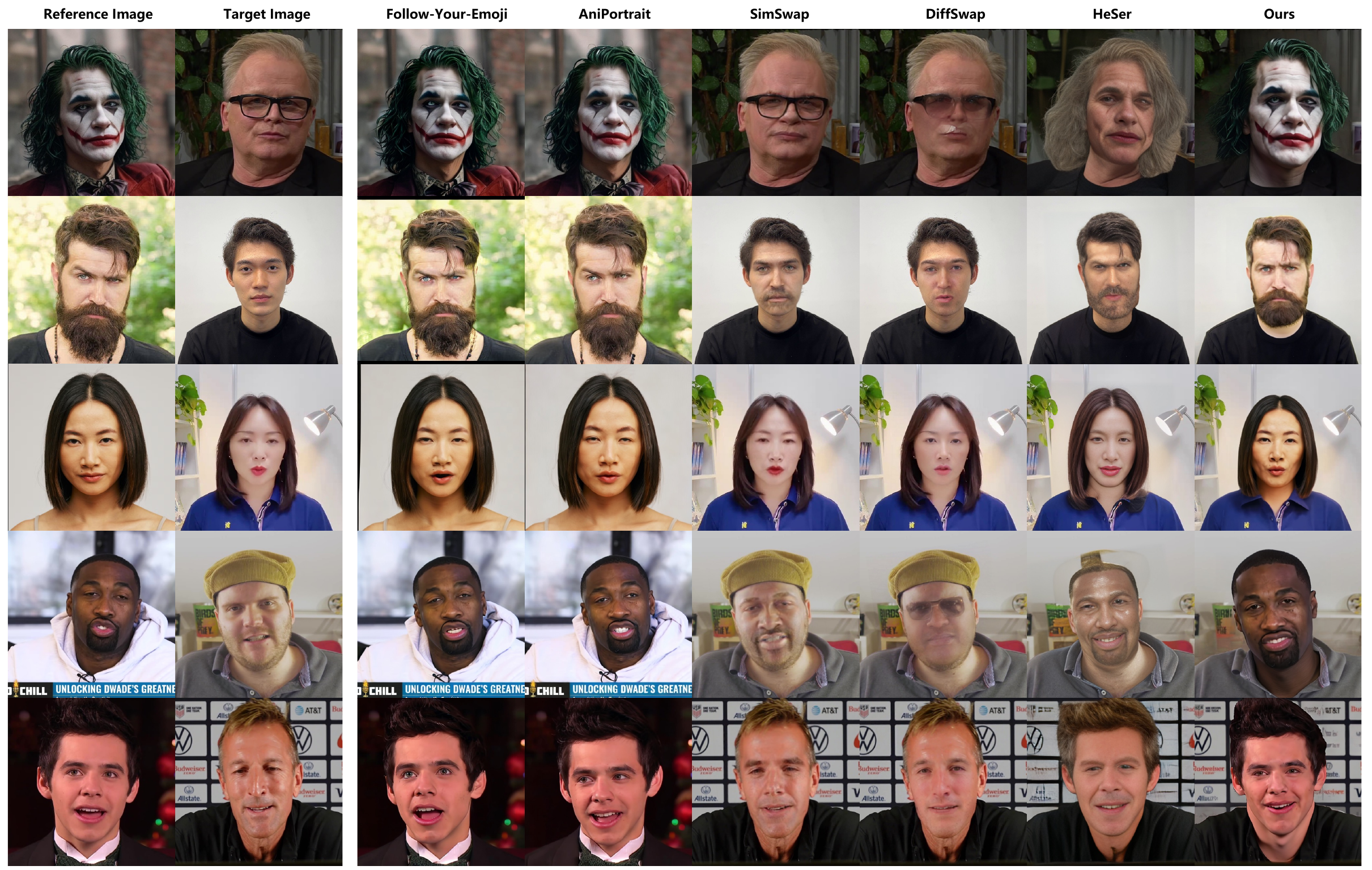}
    \caption{Qualitative comparion for head swapping.}
    \label{fig:comparsion}
\end{figure*}

To achieve high-quality head-swapping results, effective and expressive expression transfer is essential. AniPortrait \cite{aniportrait} employs a 3DMM to fit the reference image, efficiently decoupling identity, pose and expression, and reducing the reference image's expression influence. However, the expressiveness of the final result is constrained by the 3DMM's ability. In contrast, Follow-your-emoji \cite{emoji} leverages 3D landmarks detected by Mediapipe \cite{mediapipe}, offering enhanced expressiveness. Nevertheless, its landmarks retargeting strategy does not address residual expressions in the reference image and struggles with cross-identity expression transfer when there are significant differences in facial feature proportions. To overcome these challenges, we propose an innovative Expression-Aware Landmark Retargeting strategy. The pipeline is shown in Fig.\ref{fig:method_expression}. Given a reference image $I_{ref}$ and a driving sequence $I_{d}$, the \textbf{Neu-Exp Module} first utilizes the Mediapipe detector to obtain reference landmarks $L_{ref}$ and driving landmarks $L_{d}$, respectively. Then it fits the reference image using a 3DMM template \cite{faceverse} and projects the template to obtain the landmarks $L_{ref}^{t}$. Next, we set the expression coefficients of the template to 0 and reproject it to acquire the landmarks $L_{ref}^{t, exp=0}$. The neutral landmarks of reference image  $L_{ref}^{neu}$ can be represented as:
\begin{equation}
    \label{equ:exp}
    \begin{aligned}
        L_{ref}^{neu} = L_{ref}-L_{ref}^{t}+L_{ref}^{t, exp=0}
    \end{aligned}
\end{equation}
 Subsequently, a frame depicting a neutral expression is selected from the driving sequence and undergoes the same processing to produce neutral landmarks $L_{d}^{neu}$ of the driving sequence.

 After obtaining neutral landmarks for both reference and driving images, we can further adjust the expression scale of various facial features through \textbf{Scale-aware Retargeting}, such as the eyes and mouth. Taking eyes as a case study, given the top and bottom landmark indices for the eyes, denoted as $i_{eye}^{top}, i_{eye}^{bot}$, we compute the expression magnitude adjustment ratio $s_{eye}$ along yaw axis as follows:
\begin{equation}
    \begin{aligned}
        s_{eye} = \frac{L_{ref}^{neu}[i_{eye}^{bot},1] - L_{ref}^{neu}[i_{eye}^{top},1]}{L_{d}^{neu}[i_{eye}^{bot},1] - L_{d}^{neu}[i_{eye}^{top},1]} 
    \end{aligned}
\end{equation}
A similar procedure is applied to the mouth to obtain $s_{mouth}$. The scale-aware retargeting process can be formulated as:

\begin{equation}
    \begin{aligned}
        \Delta L = L_{d} - L_{d}^{neu} 
    \end{aligned}
\end{equation}

\begin{equation}
    \begin{aligned}
        L &= L_{ref}^{neu} + \Delta L + (s_{mouth} - 1) \odot \Delta L[:,i_{mouth}] \\&\quad + (s_{eye}-1) \odot \Delta L[:,i_{eye}]
    \end{aligned}
\end{equation}

where $L$ is the retargeted driving landmarks and $i_{mouth}, i_{eye}$ is the landmarks indices of mouth and eyes, respectively. Benefiting from the disentanglement of identity, expression, and pose, as well as the control of expression magnitude, our method also enables expression editing. For more details, please refer to the supplementary material.

\section{Experiment}
\label{sec:experiment}

\subsection{Implementation Details}
We train our model on the HDTF \cite{DBLP:conf/cvpr/ZhangL0F21}, Voxceleb \cite{DBLP:conf/interspeech/NagraniCZ17}, VFHQ \cite{DBLP:conf/cvpr/XieWZDS22} and TalkingHead1KH \cite{DBLP:conf/cvpr/LiLZQWJ22}. We first detect and crop the video's head region, then split the video into clips with durations of less than 30 seconds. To ensure quality, we utilize Image Quality Assessment (IQA) \cite{chen2024topiq} to filter out clips with inferior image quality. Finally, our training dataset consists of 30K video clips from about 20K identities. During the training phase, we randomly sample two frames with different expressions from the video clips (one as the reference image and the other as the target image) and the images are resized as $512 \times 512$. We employ Mediapipe \cite{mediapipe} detector to extract 3D landmarks from the target image and use lama as driving signals \cite{DBLP:conf/wacv/SuvorovLMRASKGP22} to inpaint the background. The training details are explained in the supplementary materials.

\subsection{Comparison with baselines}

\subsubsection{Qualitative results}
We compare our method with previous state-of-the-art face swapping methods SimSwap \cite{simswap}, DiffSwap \cite{diffswap}, head swapping method HeSer \cite{heser} and portrait animation methods Follow-your-emoji \cite{emoji} and AniPortrait. HeSer utilizes third-party reproduced code as the official implementation is not available. HS-Diffusion is excluded as no public code is available. The results are presented in Fig.\ref{fig:comparsion}. Face swapping methods \cite{diffswap,simswap} often fail to effectively transfer head shapes and hairstyles, resulting in poor identity consistency. DiffSwap \cite{diffswap} generates blurry images. Head swapping method \cite{heser} transfers the entire head but may introduce background artifacts, inaccurate expression transfer and identity alteration. In contrast, our method maintains identity consistency and seamless foreground-background integration, accurately transferring target image expressions, even for uncommon portrait such as the Joker at the first row. For comparison results of portrait animation , please refer to the supplementary materials.

\begin{table}[t]
\centering
\begin{tabular}{cccc}
\hline
 Method      & ID Similarity $\uparrow$ & Pose $\downarrow$   & Expression $\downarrow$   \\ \hline
Aniportrait\textsuperscript{1} & 0.892        & 6.55  & 0.025   \\
Emoji\textsuperscript{1}       & \textcolor{red}{0.906}        & 16.8 & 0.026    \\
SimSwap\textsuperscript{2}     & 0.592        & \textcolor{red}{0.702}  & \textcolor{red}{0.005}   \\
DiffSwap\textsuperscript{2}    & 0.397        & \textcolor{blue}{1.144}  & \textcolor{blue}{0.008}   \\
HeSer\textsuperscript{3}       & 0.319        & 6.840  & 0.022  \\ \hline
Ours        & \textcolor{blue}{0.895}        & 9.83  & 0.014  \\ \hline
\end{tabular}
\caption{The quantitative results. \textsuperscript{1} denotes portrait animation methods, \textsuperscript{2} denotes face swapping methods, and \textsuperscript{3} indicates head swapping methods. The optimal results are marked in red, the second-best results are marked in blue. Emoji denotes Follow-Your-Emoji.}
\label{tab:quant}
\end{table}

\subsubsection{Quantitative results}
For quantitative evaluation, we randomly selecte 200 videos with unique identities from the VFHQ \cite{DBLP:conf/cvpr/XieWZDS22} dataset, extracting 48 frames from each video at intervals of 2 as target videos. Additionally, we chose 200 portrait images from VFHQ as reference images, ensuring different identities from the choosed videos. We evaluate ID Similarity, pose error, expression error, and FID. ID similarity is measured by extracting indetity features with ArcFace \cite{arcface} and calculating cosine similarity from both generated and reference images. Pose error is assessed using a pretrained head pose estimator \cite{DBLP:conf/cvpr/RuizCR18} to extract head poses from both generated results and target videos, followed by L2 distance calculation. Expression error is evaluated by extracting expression coefficients with a facial blendshape extractor \cite{mediapipe} from both sets and computing the L2 distance. 

The results are shown in Tab.\ref{tab:quant}. For expression, our method achieves smaller errors compared to analogous method Follow-your-Emoji \cite{emoji} and AniPortrait \cite{aniportrait}. For pose error, our method is comparable to HeSer and AniPortrait, with differences likely attributable to the variations in the 3D landmarks employed. Our approach outperforms Follow-Your-Emoji that also utilizes Mediapipe landmarks. Face-swapping methods obatin smallest errors as they tend to make subtle modifications to the target image, whereas our approach involves fully regenerating the head. For ID similarity, our method substantially surpasses traditional face-swapping and head-swapping approaches by comprehensively preserving head features. While head-driven methods inherently exhibit superior ID retention without the need to integrate new background and body integration, our approach achieves comparable accuracy under these conditions. This highlights our method's robust capability for identity preservation in head-swapping scenarios.

\subsection{Ablation study}

\begin{figure}[t]
    \centering
    \includegraphics[width=\columnwidth]{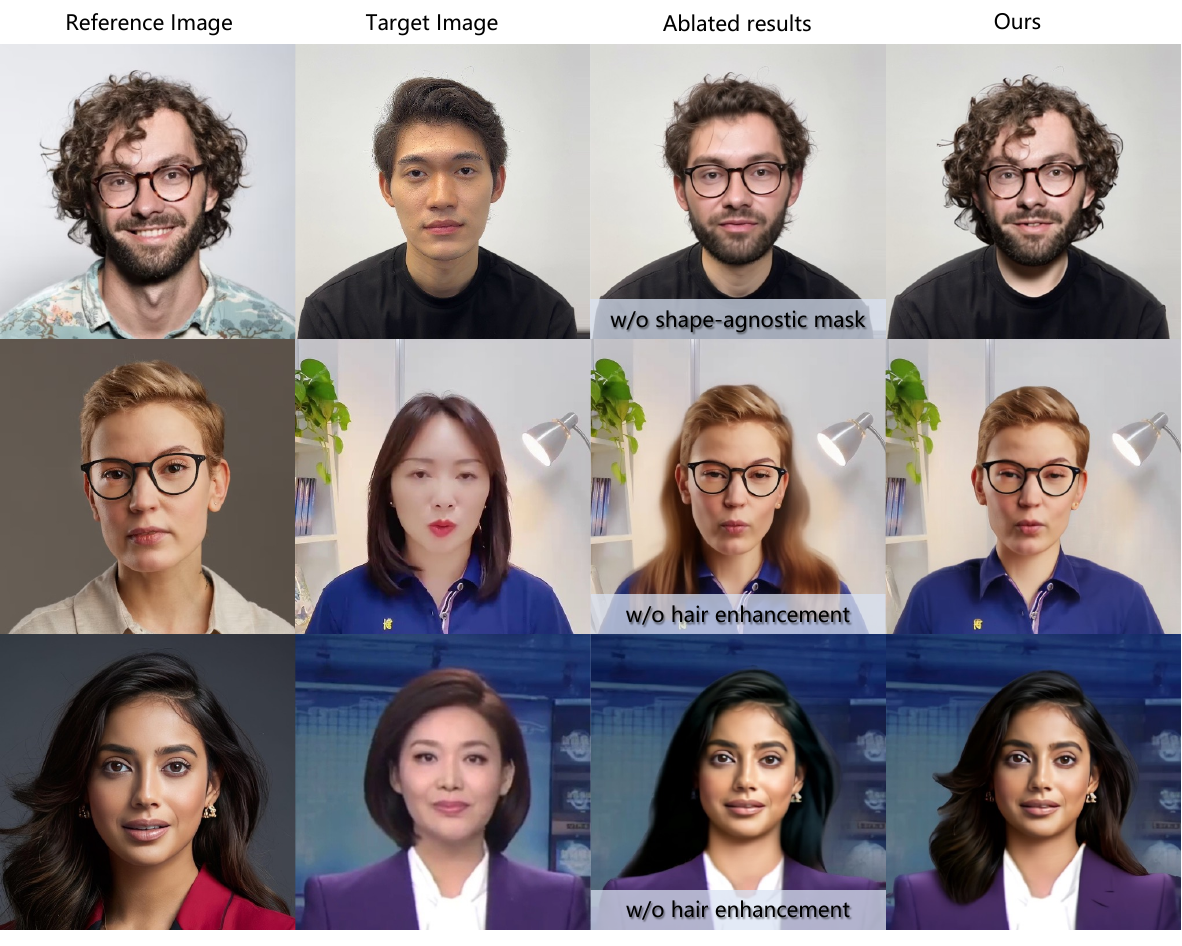}
    \caption{The ablated results of Shape-Agnostic Mask and Hair Enhancement Strategy.}
    \label{fig:ablation_shape}
\end{figure}

\begin{figure}[t]
    \centering
    \includegraphics[width=\columnwidth]{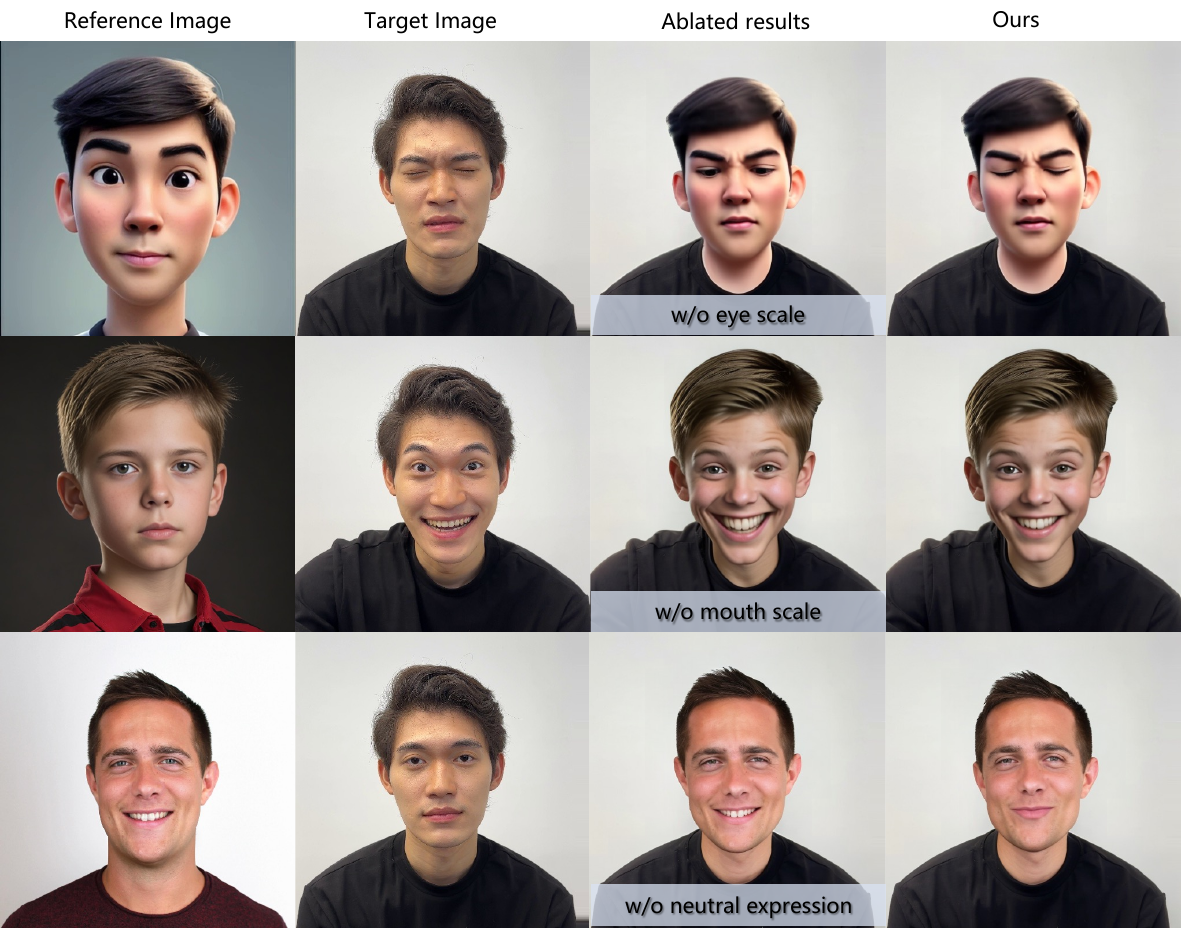}
    \caption{The ablated results of Expression-Aware Landmark Retargeting.}
    \label{fig:ablation_exp}
\end{figure}

In this section, we analyze the effectiveness of the Shape-Agnostic Mask and Hair Enhancement strategy, as well as the Expression-Aware Landmark Retargeting module.

\noindent\textbf{Effectiveness of the Shape-Agnostic Mask and Hair Enhancement.} To assess the effectiveness of the two strategies, we conduct experiments by disabling each one separately during training. The comparative results are illustrated in Fig.\ref{fig:ablation_shape}. When Shape-Agnostic Mask strategy is omitted, the generated head region is disproportionately influenced by the head shape of the original driving video, leading to head shape mismatch with the reference image. Moreover, the generated head is strictly confined within the head boundaries of the driving video, restricting the region available for hair generation. Without Hair Enhancement strategy, 
the model generates undesirable hair artifacts due to incorrect shoulder-related priors as shown in the second row. It also fails to naturally generate hair in front of the shoulders without shoulder mask, as shown in the third row. By contrast, our model maintains better identity consistency and is capable of generating diverse hairstyles.

\begin{table}[t]
\centering
\begin{tabular}{cc}
\hline
                     & Expression $\downarrow$ \\ \hline
ours(w/o Neu-Exp Module) & 0.025      \\
ours(w/o Scale-aware Retargeting)      & 0.015      \\
ours                 & \textbf{0.014}      \\ \hline
\end{tabular}
\caption{The quantitative results of the Expression-Aware Landmark Retargeting.}
\label{tab:ablation}
\end{table}

\noindent\textbf{Effectiveness of Expression-Aware Landmark Retargeting.} 
To verify the effectiveness of the Expression-Aware Landmark Retargeting module, we conduct experiments by discarding Neu-Exp Module and Scale-aware Retargeting repectively during inference. The Qualitative results are presented in Fig.\ref{fig:ablation_exp}. In the absence of Neu-Exp Module, the driven expressions overlay on the existing expressions of the reference image (for example, a smile), which reduces the accuracy of expression transfer. Furthermore, without Scale-aware Retargeting, the accuracy of expression transfer diminishes, especially between subjects with significant differences in facial feature sizes; For instance,  Mismatched eye sizes may result in improper eye closure, as shown in the first row. Variations in mouth size between the reference image and driving video can lead to exaggerated facial expressions as shown in the second row.The quantitative comparison is detailed in Tab.\ref{tab:ablation}

\section{Conclusion}
we present a novel diffusion-based multi-condition controllable framework for video head swapping that ensures background and identity consistency while enabling expressive and flexible facial expression transfer and editing. By integrating innovative strategies such as shape-agnostic masking and a novel hair enhancement strategy, our method reduces identity leakage and supports the generation of diverse hairstyles. The expression-aware landmark retargeting module enhances the accuracy and expressiveness of cross-identity expression transfer, and supports expression transfer. Our framework achieves seamless head swapping across various hairstyles and complex backgrounds, and precise expression transfer. Experimental results confirm the method's superiority on a variety of scenes.
{
    \small
    \bibliographystyle{ieeenat_fullname}
    \bibliography{main}
}

\end{document}